\documentclass{article}

\usepackage[style=apa,natbib=true]{biblatex}
\usepackage{tabulary}
\usepackage{url}
\usepackage{graphicx}
\usepackage[utf8]{inputenc}
\title{Evolved Open-Endedness in Cultural Evolution: A New Dimension in Open-Ended Evolution Research}

\author{James M. Borg$^{1*}$, Andrew Buskell$^{2}$, Rohan Kapitany $^{3,4}$\\ Simon T. Powers$^{5}$, Eva Reindl$^{6,7}$, Claudio Tennie$^{8}$
}

 \date{}
 
\begin{document}
\maketitle

\begin{center}
$^1$School of Informatics and Digital Engineering, Aston University, UK \\
$^*$Corresponding Author: j.borg@aston.ac.uk \\
$^2$Department of History and Philosophy of Science, University of Cambridge, UK \\
$^3$School of Psychology, Keele University, UK \\
$^4$School of Anthropology and Museum Ethnography, University of Oxford, UK\\
$^5$School of Computing, Edinburgh Napier University, UK \\
$^6$Department of Anthropology, Durham University, UK \\
$^7$School of Psychology and Neuroscience, University of St Andrews, UK \\
$^8$Department of Early Prehistory and Quaternary Ecology, University of T{\"u}bingen, Germany \\
\end{center}

\begin{abstract}
The goal of Artificial Life research, as articulated by Chris Langton, is ``to contribute to theoretical biology by locating life-as-we-know-it within the larger picture of life-as-it-could-be'' (\citeyear[p.~1]{langton1989artificial}). The study and pursuit of open-ended evolution in artificial evolutionary systems exemplify this goal. However, open-ended evolution research is hampered by two fundamental issues: the struggle to replicate open-endedness in an artificial evolutionary system, and the fact that we only have one system (genetic evolution) from which to draw inspiration. We argue that cultural evolution should be seen not only as another real-world example of an open-ended evolutionary system, but that the unique qualities seen in cultural evolution provide us with a new perspective from which we can assess the fundamental properties of, and ask new questions about, open-ended evolutionary systems, especially in regard to evolved open-endedness and transitions from bounded to unbounded evolution. Here we provide an overview of culture as an evolutionary system, highlight the interesting case of human cultural evolution as an open-ended evolutionary system, and contextualise cultural evolution by developing a new framework of (evolved) open-ended evolution. We go on to provide a set of new questions that can be asked once we consider cultural evolution within the framework of open-ended evolution, and introduce new insights that we may be able to gain about evolved open-endedness as a result of asking these questions.

\end{abstract}



\section{Introduction}

Genetic evolution appears to be open-ended. Taking advantage of environmental regularities, gene expression and regulation can generate a potentially infinite number of traits and trait variations. Such evolutionary open-endedness has been characterized by a constellation of overlapping features, yet can generally be understood as the ability of an evolutionary system to produce a continuous stream of novel units \citep{taylor2016open}. For those trying to create and understand open-ended evolutionary systems the goal is to understand the underlying principles and dynamics of evolutionary systems in general. Such understanding is based upon knowledge of the best explored and understood open-ended evolutionary system: genetic evolution. But it also can, and should,  draw upon the development of artificial evolutionary systems that  explore the principles of life-as-it-could-be \citep{langton1989artificial}. Such artificial evolutionary systems depart from the rules and principles of Darwinian genetic evolution while still meeting the general requirements of an evolving system. The interaction between the two can be consilient. Darwinian genetic evolution provides a source of valuable ideas and inspiration as well as justification for the designs of artificial systems. Despite this positive interplay, having only one concrete instance of an open-ended system is a problem. Such sparse epistemological situations can limit abilities to discern alternate possibilities, detect generalizable features, and develop robust theories and models. 

It is increasingly being recognised, however, that there is another evolutionary system from which one can find inspiration: cultural evolution \citep{marriott2018social,borg2021evolved, bedau2013minimal, bedau2019open, bedau2019patented}. Minimally characterized, culture is information transmitted through mechanisms of social learning \citep{boyd1985culture, CES, whiten2022emergence}. And while this minimal characterization leaves out many distinctive features of human and non-human cultural groups (as e.g. different species differ in the types of information they can transmit) \citep{whiten2022emergence,buskell2019systems}, and leaves open precisely how `social learning' should be construed \citep{lewens2015culture}, its abstractness makes it exceptionally useful for designing models of cultural change and describing general evolutionary dynamics. On this characterization, cultural evolution is the change in frequency -- or, of special interest here -- the form of cultural traits over time, where these changes are at least in part influenced by social learning \citep{neadle2017food}. 
Although cultural evolution is often described as being analogous to genetic evolution \citep{cavalli1981cultural}, there are clear differences in the way culture is inherited: 1. while genetic evolution relies on typically two (sometimes one) parent(s), there are potentially unlimited numbers of cultural ``parents''; 2. while genetic transmission is almost exclusively transmitted vertically from parent to child, cultural transmission can involve substantial amounts of horizontal or oblique transmission; 3. while genetic changes generally occur between generations, cultural change generally occurs within generations \citep{mesoudi2011cultevol,mesoudi2006unified}. While these features distinguish cultural from genetic change, these do not imply that cultural inheritance is in any sense less (or not) ``evolutionary'' -- only that its dynamics frequently differ.

Over the past 40 years there has been increasing recognition that culture and cultural evolution exist within non-human animal populations (most prominently in birds and mammals) \citep{whiten2021burgeoning,whiten2021psychological,whiten2019cultevol}, and that culture not only exists as a result of genetic adaptation but also plays an important co-evolutionary role in guiding genetic evolution \citep{whitehead2019reach,uchiyama2021}. This co-evolutionary relationship between genes, culture, and the environment is sometimes known as ``triple inheritance" \citep{laland2000triple}. Nonetheless, while many animal species exhibit culture, human cultural evolution appears both quantitatively and qualitatively distinct. Several dividing lines between human and animal cultures have been proposed, but the most prominent of recent formulations holds that human culture is distinctive in virtue of its cumulative nature -- with human culture accumulating modifications over time, and with these modifications building upon one another \citep{tomasello1999origins}. However, as more observations of cultural evolution in other species have been made, it has become increasingly apparent that cumulative cultural evolution is actually not unique to human culture \citep{mesoudi2018cumulative}. This raises the following question: what, if anything, is unique about human cultural evolution?

We think issues about the distinctiveness of human culture and the nature of open-ended evolution are overlapping -- and that explorations of the two will be mutually illuminating, with potential downstream consequences for Artificial Life. Here we situate cultural evolution within a broader framework of open-ended evolution and argue that: 
\begin{enumerate}
    \item Culture is an evolving system, co-evolving alongside genetic evolution. 
    \item That within cultural species there are a range of ``types'' of cultural evolutionary patterns; cumulative and non-cumulative, tall and wide, unbounded and bounded.
    \item That recognizing these ``types'' of cultural evolution allows Artificial Life researchers to better understand evolutionary dynamics and provides new perspectives from which to explore open-ended evolution.
    \item That only humans demonstrate open-ended cultural evolution and that human cultural evolution has transitioned from a bounded to an unbounded evolutionary system in recent evolutionary history, thus providing a second instance of ``evolved open-endedness.''
    \item That existing Artificial Life methods can be fruitfully applied to the study of cultural evolution.
\end{enumerate} 

To develop these points, we outline a number of core concepts from the wider study of cultural evolution. We then analyze ``open-ended evolution'' and explore how such analyses might improve our understanding of evolutionary dynamics and the emergence of evolved open-ended evolutionary systems. A table of definitions for the key terms used here can be found in table \ref{tab:definitions}.

\begin{table}[h]
\centering
\begin{tabulary}{1.0\textwidth}{LLL}
Term & Definition & See\\
\hline
\hline
Culture & Information transmitted through mechanisms of social learning  & \citet{boyd1985culture, CES, whiten2022emergence} \\
\hline
Cultural Evolution & The change in frequency or the form of cultural traits over time, where these changes are at least in part influenced by social learning  & \citet{neadle2017food} \\
\hline
Open-Ended Evolution & An evolutionary process that is capable of producing a continuous stream of new adaptive novel units, with no \emph{a priori} limitations on the generation of such novelty & \citet{taylor2016open, gabora2017autocatalytic} \\
\hline
Cumulative Culture & A process whereby a culturally transmitted trait accumulates modifications over time with a ratchet-like effect & \citet{tomasello1999origins,boyd1985culture} \\
\hline
Unbounded Evolution & A continuous demonstration of new adaptive novelty and/or the ongoing growth in trait diversity. Term used interchangeably with open-ended evolution, but often used to contrast with bounded evolution & \citet{bedau1998classification, channon2006unbounded} \\
\hline
Evolved Open-Endedness & Open-endedness as the outcome of an evolutionary process as opposed to an assumed pre-condition & \citet{pattee2019evolved}\\
\hline
Wide Evolution & A characterization of the disparity of traits and traditions; increased through processes of recombination, innovation, or the exploration of previously underappreciated affordances & \citet{buskellForthcomplex, derex2022exploit} \\
\hline
Tall Evolution & A characterization of the typical length (measured in relevant changes generated through cumulative evolution) of independent trait traditions & 
\end{tabulary}
\caption{Reference table of definitions for key terms}
\label{tab:definitions}
\end{table}

\section{Cultural Evolution}
What is culture and how does it evolve? As suggested above, culture can be minimally defined as the transmission of information -- traits -- through mechanisms of social learning \citep{boyd1985culture}. This minimal and abstract characterization of culture permits ``information" and ``traits" to be read in an encompassing way to include a wide variety of techniques, technology, and behavior.  Examples of such traits include the extractive foraging techniques among chimpanzees \citep{sanz2010chimpanzees} or methods for lighting a fire \citep{macdonald2021middle}. It may also incorporate behaviors with communicative effects such as warning calls \citep{griffin2004social}, bird-song, or language \citep{janik2000different}. The definition also incorporates population-level conventions among conspecifics for greeting and leave-taking \citep{duranti1997universal,baehren2022saying} as well as normative behaviors such as styles of dress or decoration \citep{baehren2022saying, richerson2009tribal}. Again, the key is that the acquisition of these behavioral traits or beliefs are and must be influenced by social learning -- when they are not, the traits are not cultural. 

\subsection{Does Culture Evolve?}
An evolutionary process does not require a \emph{particular} kind of physical instantiation or biological substrate. While familiar processes of biological evolution are mainly grounded in the manipulation and modification of genes, cultural evolution (and evolution more generally) is under no such obligation. Consider Dennett’s \citeyear{dennett1996darwin} conception of evolution as being both algorithmic and substrate neutral. Evolution is algorithmic in the sense that if certain conditions are met, a certain sort of outcome is necessarily produced \citep[p.~48]{dennett1996darwin}. Where there is reproduction with variation under selection at a population level, a certain kind of outcome is produced -- in this case, the frequency of adaptive outcomes is increased in the population over time. In cultural evolution, `adaptive' may refer to the cultural trait and the success the trait has in spreading from mind to mind \citep{rosenberg2017social}, or it may refer to the effects the trait has on its bearers' (adaptive) behavior, though the co-evolutionary nature of culture and biology, and the effect culture has on biology, cannot (and should not) be understated \citep{henrich2007dual}. 

Ultimately, the target of reproduction is the informational content carried by some vehicle -- whether this vehicle is expressed behavior, an artefact, or the instructions of a written account (though it is of course the case that the vehicle can itself have ``fitness''). Artificial Life has often equated such a characterization with the idea of a ``meme'' \citep{bull2000meme, bullinaria2010memes, bedau2013minimal}: a discrete, particulate unit of information that is copied intact between brains, analogous to the way that genes are copied between parents and offspring \citep{dawkins1976selfish}. Cultural evolution, however, does not require the process of reproduction and cultural inheritance to be understood in terms of strict copying. While the literature on this point is vast, \citet{rosenberg2017social} provides a clear summary of the arguments: 
\begin{enumerate}
    \item Replication in biology has not always involved high-fidelity replicators -- the ``major transitions in evolution'' literature explains how evolution itself has gradually generated higher fidelity transmission processes. While the first replicating molecules were not DNA, nor did they have accurate copying mechanisms, fidelity increases are evolutionary achievements that could and can be selected for over time \citep{maynardsmith1995major}.
    \item Even in genetic evolution, a single gene can rarely be equated with a single trait -- the vast majority of biological traits result from complex interactions between the proteins expressed and regulated by many genes, so why should one demand in cultural evolution that a trait is the product of one discrete meme?
    \item Many features of human institutions are adapted to preserve and proliferate cultural traits even under low individual copying fidelity. Variation is introduced in the form of the (re)combination of existing traits, innovation of new traits by individuals (which may involve rational thought), or copying error (loosely analogous to mutation in genetic evolution). Meanwhile, selection may occur in multiple ways. This includes biological selection -- that is, the effect that cultural traits have on biological fitness (for instance, being led to believe that something is safe to eat when it is not).
\end{enumerate}
If we accept that evolution is algorithmic (i.e. it follows a series of processes to produce a certain outcome; selection + reproduction + variation = evolution), it follows that we are not bound to particular features of biological processes (e.g. sexual reproduction), nor are we bound to a specific substrate (e.g. DNA). Though Dennett’s conception of cultural evolution is considered out-of-date to some modern scholars of cultural evolution \citep{uhlivr2012needs}, his fundamental argument applies to it nonetheless: the idea of an algorithmic process makes it all the more powerful, since the substrate neutrality it thereby possesses permits us to consider its applications to just about anything \citep{dennett1996darwin}. Which is, of course, true: That one can create an evolutionary process within a computer is evidence that the process itself need not be strictly biological, merely algorithmic \citep{lehman2020surprising}. 

\subsection{Co-Dependent Evolutionary Systems}
Cultural evolution is deeply intertwined with biological evolution. While these evolutionary processes and their products can generate complicated co-evolutionary feedback loops, each evolutionary system can be understood, studied, and modelled separately \citep{boyd1985culture, mesoudi2011cultevol}. For instance, as we suggest in more detail below, pre-modern hominin cultural evolution contributed to biological fitness in the form of ecological knowledge and technological production. Nonetheless, over time, cultural evolution has become increasingly unmoored from genetic fitness effects, producing a wide range of behavioral, social, and technological change \citep{henrich2015secret}. The reason for both the intimacy and relative independence of the two systems should be evident. The substrate of culture is biological: the brain. 

Culture is bound to a biological substrate, but a substrate which is different from the classical understanding of genetic evolution in which traits are encoded (directly or indirectly) by genes. Gene expression may produce brains and (some) brains may acquire culture, but one cannot skip the middle step and claim that genes produce culture. While humans may be biologically prepared to acquire language \citep{fitch2011unity}, they are not biologically determined to learn English, Farsi, or Korean. Clearly, accessibility and exposure to certain kinds of inputs -- the presence of English, Farsi, or Korean language cues -- determine what language any given human ultimately produces. Or put another way, the acquisition, production, and transmission of language is largely influenced by social learning. So one cannot simply claim that the process of cultural evolution is independent from biology. Biological and cultural evolution are interdependent. 

The idea that cultural species, and particularly cumulatively cultural species such as \emph{Homo sapiens}, have two interdependent systems of inheritance has been labelled `dual inheritance' (`triple inheritance' if the environment is also included \citep{laland2000triple}). On this account, human offspring inherit a genotype from their parents through sexual reproduction and they inherit a body of cultural information over the course of their post-natal lives via processes of social learning \citep{henrich2007dual} -- processes that themselves may be culturally evolved tools \citep{heyes2018gadgets}. Just as one’s genotype has been dictated by a history of selection pressures acting on genetic variation, one’s cultural inheritance is similarly shaped by selective pressures and the variation introduced through innovation, recombination, and error involved in social learning. Thus, in the same way that certain phenotypic features are adaptations -- increasing the biological fitness of individuals -- elements of culture may also be adaptations. Consider food taboos present in Fijian society \citep{henrich2010evolution,mckerracher2016food} which apply exclusively to pregnant women. Despite the causal opacity of the underlying process, these taboos protect women from miscarriage. Alternatively, consider the ritualized process of cassava production. Again, despite the causal opacity of the underlying process, populations have developed practices that remove toxic cyanogenic elements which would have long-term health consequences if regularly consumed \citep{bradbury2011mild,cardoso2005processing,mckerracher2016food,banea1992shortcuts}. Of course, it can also be adaptive to acquire cultural elements idiosyncratic to local cultures. Regardless of whether the practice of female or male circumcision has biological benefits, within a circumcising culture, it can be adaptive to demonstrate commitment to the group by engaging in such a costly signal. This can ensure inclusion and support by the group as well as prevent ostracism \citep{sosis2004adaptive, howard2017frequency} -- thus enhancing reproductive outcomes. 


Cultural organisms do not only inherit genes and cultural information, but also an environment: that is, a habitat that has been selected, modified, and partly created by their ancestors. All organisms change their habitats through their actions -- of which spiderwebs, termite mounds, or human-made earthworks are just a few notable examples -- with more or less transitory effects. Such organism-modified environments are evolutionarily relevant insofar as they modify selection pressures or transmission opportunities -- what the evolutionary literature calls niche construction \citep{laland2000triple}. Systematic and long-lasting modifications, such as beaver dam-building or human agriculture can have profound effects on both biological and cultural evolutionary processes of the species producing these modifications as well as others in the habitat.
 
While niche construction is not uniquely human, humans are distinctive in that most of their niche construction activities are cultural (e.g., making dams, fences, bridges, schools, roads, clothes). Over evolutionary time, the hominin lineage has created a cultural niche that has not only affected their biological and cultural evolution by creating new selection pressures, but which has increasingly become crucial for their survival \citep{laland2011cultural,uchiyama2021}. For example, the use of fire and cooking may have facilitated selection for larger brains alongside smaller guts and jaws. Lacking fire or cooking, hominins would have been poorly adapted to their environments \citep{aiello1995expensive}. The second inheritance system -- culture -- can thus indirectly affect the first -- genes -- through niche construction. Genes and culture have co-evolved: cultural activities such as tool use and tool making have generated selection pressures for social tolerance and cognitive skills such as social learning, attention, working memory, and language, which in turn have opened up ever greater capacities for cultural innovations, social learning, and large-scale cooperation \citep{henrich2015secret}, creating the biological and cultural conditions for the emergence of open-ended cultural evolution.

Cultural evolution is often faster than genetic evolution: a cultural variant can emerge and recombine quickly and repeatedly within the lifetime of its carrier, and can die independently of the death of the individual \citep{boyd2013cultural}. Alongside the speed of cultural evolution, humans’ capacity for planning and foresight suggests that many human adaptations are cultural or have cultural origins \citep{uchiyama2021}. Thus, cultural evolution cannot only produce solutions to (ecological) problems, but also create new opportunities and niches that cultural evolution can exploit - an autocatalytic process, resulting in the emergence of unbounded cumulative culture.

\section{Open-Ended Cultural Evolution}
As noted in the introduction, open-ended evolution is an umbrella term for a constellation of features associated with evolutionary change. These include the ongoing generation of novelties, adaptations, and evolutionary salient entities \citep{taylor2016open}. For simplicity, we hold that an evolutionary system can generate open-ended evolutionary change if it is able to produce a continuous stream of novel units (evolutionary individuals, traits) with no \emph{a priori} limits to the generation of such novelties \citep{gabora2017autocatalytic,taylor2016open}. As several commentators have noted \citep{pattee2019evolved,tennie2018culture,bedau2019open,bedau2019patented}, human cultural evolution appears to be just such an open-ended evolutionary system.

More recently, cultural evolution researchers have used the term “open-ended” to describe what is unique about human culture \citep{tennie2018culture}. This acknowledges that human culture frequently involves processes of cumulative cultural evolution -- processes that generate traits (e.g. behaviour, beliefs) that build upon previous traits, perhaps also making them more complex, efficient, and adaptive. But calling human culture ``open-ended" is also meant to suggest that cultural solutions to problems do not need to be stuck at local optima, but can break free and further improve, for instance, by the harnessing of new affordances \citep{arthur2009nature,derex2022exploit}. Focusing on this putative “uniqueness” of human culture, researchers have identified important transitions, cognitive capacities, and patterns of cultural evolution as hominins have evolved and changed over the past 8 million years. 

In the next three subsections we make distinctions between patterns of cultural evolutionary change: between cumulative and non-cumulative cultural traditions; between ``building-up'' or \emph{tall} traditions and the ``building-out'' of \emph{wide} repertoires of traditions; and between bounded and unbounded evolution. These  patterns capture important differences in cultural evolutionary dynamics. Though these patterns are distinct, they likely overlap in any given instance. In the final subsection we turn to consider how these distinct kinds of evolutionary patterns help characterize and explain the evolution of open-ended cultural evolution in hominins.

In focusing on distinct kinds of evolutionary patterns, and tracing these patterns back to concrete changes in selection pressures, cognitive mechanisms, and social arrangements, the approach taken here differs from recent attempts at describing hallmarks of open-ended evolution \citep{taylor2016open}. Hallmarks are signals, such that if one encountered them, this is good evidence that the evolutionary system is capable of open-ended evolution. By contrast, our approach distinguishes patterns that are associated with processes supporting cultural evolutionary change. These processes are critical to, but not necessarily sufficient for, open-ended evolution -- and thus are poor candidates for a hallmark approach. Nonetheless, distinguishing these processes helps to identify those important for evolving open-endedness, as well as how the interaction between such processes may be important to the eventual emergence of a system supporting full-blown open-ended evolution.

\subsection{Cumulative vs. Non-Cumulative}
A key distinction drawn by cultural evolution researchers is that between cumulative and non-cumulative culture. As many researchers see it, cumulative culture is central to explaining how human beings could have developed the sophisticated technical toolkits that allowed them to survive and thrive across varying -- and sometimes extreme -- ecologies \citep{richerson2005notbygenes, henrich2015secret, potts2013hominin,grove2011speciation}. Based on extensive human and non-human experiments, and a number of computational and mathematical models, \citet{mesoudi2018cumulative} have suggested ``core'' criteria that cultural evolutionary processes would have to satisfy in order to be classified as cumulative:
\begin{enumerate}
    \item a change in behavior, followed by ...
    \item ... transfer of the modified or novel trait via social learning, where ...
    \item ... the learned trait results in an ``improvement" in performance/fitness (cultural or genetic), with ...
    \item ... the previous steps repeated in a manner that results in (sequential) modification and improvement over time.
\end{enumerate}
However, we follow recent work in denying that ``improvement'' over time is a necessary feature of cumulative culture evolution, and instead favor a minimal formulation that sheds this requirement \citep{buskellinpressmere}.

On this minimal formulation, cumulative culture is simply the modification to, and retention of, socially transmitted cultural traits \citep{buskellinpressmere}. What we have called processes of cumulative culture in the above discussion, are whatever cognitive and social capacities are sufficient to bring about trait modification and retention over time. But these \emph{processes} generate \emph{patterns} in the evolutionary record. Because cumulative culture involves retained modifications, they have histories -- and can be considered ``traditions''. The histories of such traditions can, at least in principle, be reconstructed as sequences of step-by-step changes (akin to what \citet{calcott2009lineage} calls “lineage explanations”). This minimal formulation better aligns cumulative culture with evolution theory, such that cumulative changes can generate not only adaptive traditions, but also neutral and maladaptive ones \citep{buskellinpressmere}.

Contrasting with cumulative culture is non-cumulative cultural evolution. The latter is a process of cultural change that does not retain modifications for one reason or another. This might be because there is no retention of past behavior, no introduction of modifications, or no social learning sophisticated enough to pick up on relevant modifications. These situations might occur if individuals can only innovate new traits, cycle through a set of traits, or do not learn from one another. In these cases, histories of modifications will be non-existent, uninformative, or based in non-cultural inheritance systems. 

\subsection{Tall vs. Wide Evolution}

\begin{figure}[tp] 
    \centering
    \includegraphics[width=0.8\textwidth]{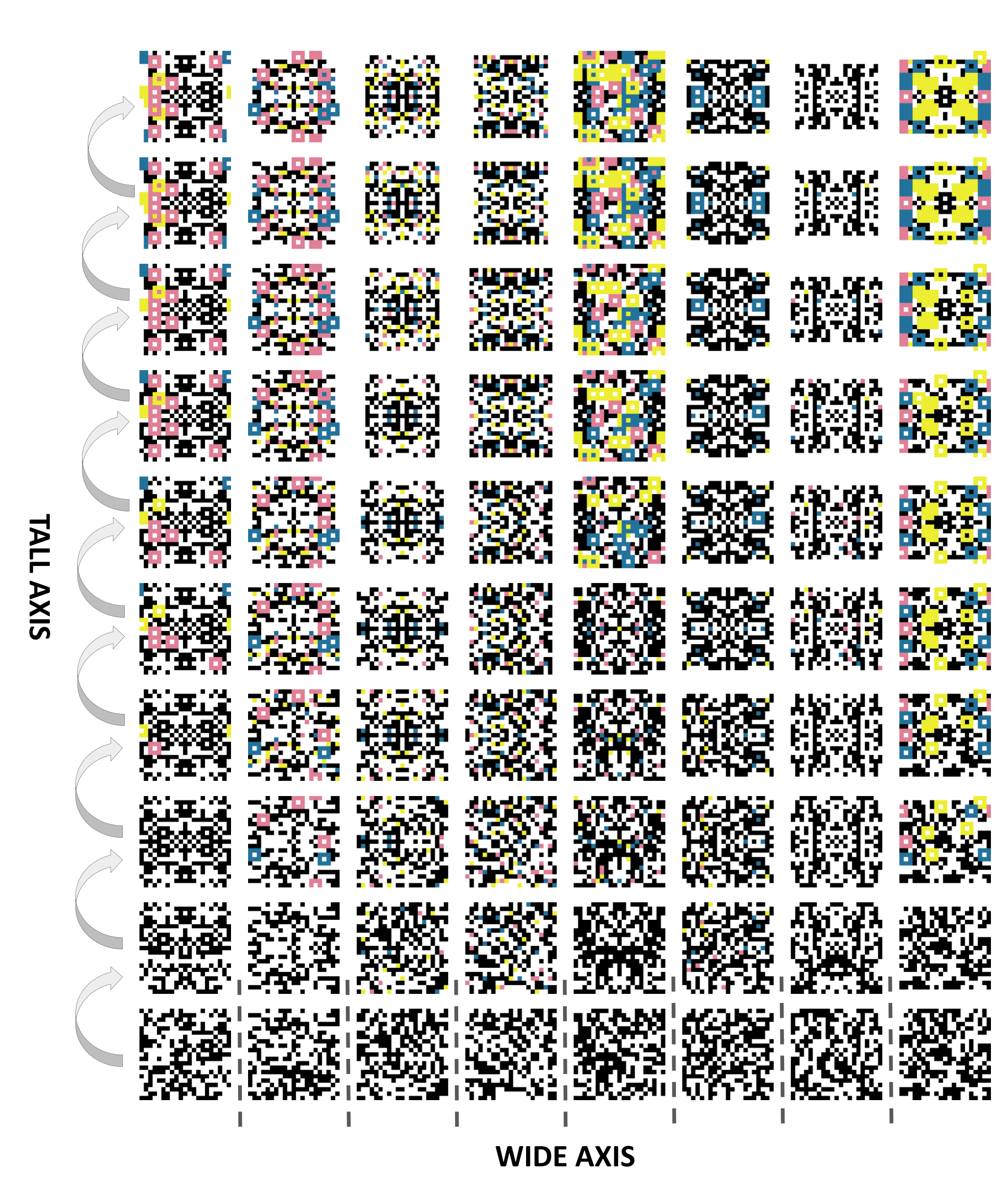}
    \caption{ This figure illustrates our conception of \textit{tall} and \textit{wide} evolution. Full details are available in SUPP MATERIAL X. Each box represents some kind of technology or cultural practice. Each pixel within each box represents a piece of discrete (but arbitrary) information. The eight squares in row 1 were generated by asking each of the pixels to become black or white at a probability of .5. Thus, all initial configurations of aggregate information are equiprobable. Thereafter one of eight arbitrary rules was applied over ten iterations. These rules were not grounded, but represent changes of `information' within the aggregate, or which introduces structure (such as symmetry) in the aggregate. As can be seen, as the aggregate information cumulatively changes over time, it becomes more complex and more structured, and increasingly dissimilar from other traditions. Each column is independent of all other columns, and `movement' along the \textit{wide} axis is not possible without violating the cumulative principle of \textit{tall} evolution.}
    \label{fig:tallvswide}
\end{figure}

Recent work has built upon analyses of cumulative culture to distinguish further cultural evolutionary patterns that had been unhelpfully lumped together. This work distinguishes between patterns involving an increasing stock of  cultural traditions (``cultural disparity'') and important aspects of cumulative cultural traditions (e.g. increases in adaptiveness, efficacy, or complexity) \citep{buskell2018religion, buskellForthcomplex}. This and other work \citep{dean2014human, tennie2009ratcheting} points to a helpful distinction between cultural evolutionary patterns: between ``building upon” traditions and ``building out” to generate new traditions -- or just \emph{tall} versus \emph{wide} evolution. 

Figure \ref{fig:tallvswide} provides a visual example of both tall and wide evolution, with tall evolution displaying a series of path-dependent adaptations within a single tradition. Each step in the sequence could only have occurred if the previous evolutionary steps had already arisen. While tall traditions need not be path-dependent -- for instance, if evolution is highly constrained -- it is a common assumption that evolutionary change is so, and we emphasize path-dependency here. Wide evolution, by contrast, is about the novel instancing of new traits. Paradigmatically, this involves the innovation of completely new traditions that need not follow any \emph{a priori} sequence. Of course, some new traditions may only arise through path-dependent cumulative evolution and recombination -- but we put those instances to the side in this illustration. Thus, in this figure, one could re-arrange the wide axis (since new traditions need not appear in any sequence), but not the tall (since each step is strongly determined by the one prior).

By way of example, let us consider some kind of adaptive problem that may have multiple starting points - starting points which are either equiprobable (equally likely to occur in the same environment), or equally efficient at solving the problem but are the product of different affordances due to different environments. This might include capturing fish, or preserving meat, or could include production of housing or clothing, refining ore into more valuable products, or skinning cats. The specifics matter less than the principle being illustrated. Along the x-axis we have multiple starting points. Let us consider the fishing example. One equiprobable starting point may be to wait in the shallows and bash a fish with a rock as it swims by, or, to bash a fish with a stick. Another example may be to wait at a certain point on the beach which, at low-tide, forms a natural pool from which fish cannot escape.  Another yet may involve poisoning the water with certain plant foliage. It can be true that these starting points are 1) are all equally likely due to the affordances of the environment, or 2) are all arrived at by different groups who live in different environments with different affordances. Whether either is true in any given situation is less important than accepting that these are (some of) the starting points for acquiring fish.

Tall evolution may involve the rock culture innovating upon the basic rock-bashing behaviour. Perhaps first by throwing the rock, then to tying a fibre to the rock before throwing (so as to recover the rock more quickly through a pulling motion); and then using multiple rock-fiber devices to expand the range of striking. Later innovations might eschew the bashing/throwing motion for connecting the fibers together to make a rake or net. Further innovating might then improve the netting technology or the casting technique, and so on. 

Meanwhile, the stick culture may innovate upon the bashing motion by innovating a sharp point -- now preferring to pierce rather than to bash. Later innovations might make spears much longer than would ever be practical for bashing, so as to stand further away from the fish without scaring them. Then, perhaps, innovations might lead to a stone-tip for the spear. And later still, a spear-throwing device like an atlatl or woomera to bring down larger prey, and so on. 

It may be the case that the first instance that stick bashing and rock bashing are equally (in)efficient, and that -- assuming an abundance of rocks and sticks -- one individual or one culture may switch between techniques with little cost. However, once groups begin to innovate upon their starting point, horizontal movement comes with greater cost, and relies upon different principles. A raking technique does not beget a spear-thrower, and vice versa. After `tall' evolution has progressed beyond a certain point, horizontal movement cannot be integrated/combined with the existing `advanced' approach, and switching comes at greater cost to the individual or culture. 

Another case study is the tool use of chimpanzees. Chimpanzees are capable of spontaneously innovating tools given available resources, such as using blades of grass for termite fishing, sticks for obtaining out of reach objects, branches for scooping algae out of water \citep{boesch1990tool,sanz2010chimpanzees,bandini2017scooping}. Each and all of these innovations can exist within a population of individuals, but the existence of one need not depend on the existence of any other. Theoretically, any of these innovations can be selected for and spread within the population independently of the others. This is wide evolution. Nonetheless, modifications could be added to these innovations -- introducing an anvil-prop to nut-cracking, chewing and stripping the grass to produce ant-catching bristles -- that put them on the vertical road to becoming a tall cultural evolutionary tradition. 

This example also points to an important corollary of the distinction between tall and wide evolution. The capacities underlying each plausibly come apart. This seems clear when one looks at hominin evolution, where early capacities for social learning lead to wide knowledge bases of disparate ecological traditions prior to the building up any particular tradition into more complex forms \citep{buskellinpressmere, sterelny2021pleistocene} (more on this below).

More generally, we want to resist identifying tall or wide evolution patterns as hallmarks of open-ended evolution. It is an open question of how tall (or short), wide (or narrow) evolutionary patterns relate to open-ended evolution, as well as the transition to open-ended evolution. As examples above and below suggest, capacities that support tall and wide evolutionary patterns likely existed well before ecological and evolutionary circumstances permitted their expression. And indeed, open-endedness most likely emerged from the gradual accumulation of new traditions, their elaboration into tall, path-dependent traditions, and their recombination and exaptation into bushy, wide, and novel traditions - we can see this visually in the patent record genealogies produce by \citet{bedau2013minimal, bedau2019patented}, with both the gradual accumulation of new patent traditions and long sequences of traditions building up being easy to identify. There's no reason to take either tall or wide evolution as a hallmark of open-ended evolution, ultimately they just describe the patterns of change that underpin the emergence of open-endeded evolutionary process, both seem necessary for open-ended evolution to emerge, but only further empirical analysis of the patterns of change found in open-ended evolutionary systems will allow us to ascertain whether common pattern exists or whether a multitude of patterns can ultimately underpin open-endedness. This should be unsurprising. Both formal modelling \citep{enquist2010one,kolodny2015evolution,winters2020comb} and cultural evolutionary theory \citep{richerson2005notbygenes,buskell2019systems,charbonneau2016modularity} emphasizes the role of cultural recombination as a potent force in generating new innovations: this occurs when distinct cultural traditions (or their constituent elements) are combined, and potentially exapted \citep{mesoudi2018cumulative}, to generate new traits. We expand upon this line of thinking below and go on to ask whether these variations in the progression of evolution (tall, wide, recombinative, exapted) are detectable within the ``ALife test'' introduced by \citet{bedau1998classification} (also see, \citet{channon2001passing,channon2003improving,channon2006unbounded}).

\subsection{Unbounded/Bounded Evolution}
A conceptually distinct and contrasting set of evolutionary patterns is that between bounded and unbounded evolution. Bounded evolution occurs when abilities for transmission, retention, or the production of modifications are limited or absent. This leads to evolutionary exploration of a parochial, bounded space of traits. Unbounded evolution, by contrast, occurs when the above abilities for transmission, retention, or the production of modifications are present and when the environment facilitates evolutionary exploration. This might occur, for instance, when the environment is rich in natural resources which can be exploited in technological production \citep{derex2022exploit}. 

To get a grasp on this distinction, it is useful to look at a domain in cultural evolutionary research where issues of boundedness or unboundedness arise. A good example is work on the Zone of Latent Solutions (ZLS) Theory \citep{tennie2009ratcheting}, which analyses the cultural and putative cumulative cultural traditions of non-human animals. \emph{Putative}, because while several species have capacities for social learning, they appear to have minimal capacities for building upon previous traits. Speaking generally, the ZLS theory suggests that the cultural capacities of non-human animal species are ``bounded", limited by a possible range of features. Explanations for why this might be the case have mainly centred on the great apes (hereafter ``apes"), but developing work suggests similar explanations may hold true with other animals, such as some birds and whales \citep{aplin2019birds,perry2011traditions,vanschaik2003orangculture,whitehead2015whales,whiten1999cultures}.

According to the ZLS theory, many putative instances of ape (and perhaps other animals') cumulative culture are not, in fact, instances of cumulative culture. The ZLS theory argues that apes lack (or have minimal, or rarely expressed) capacities for transmitting and retaining trait modifications. What appears to be cumulative culture is instead likely to be socially-influenced \emph{reinnovation}. When apes reinnovate, they draw on a baseline repertoire of behaviours -- behaviours that any able-bodied ape would be able to express -- to individually strike upon the trait of interest. Though this reinnovation may be socially facilitated, in the sense that other apes may draw attention to relevant or highly salient environments or objects, the trait is developed by each learner anew. 

The basic idea of the ZLS is that this baseline repertoire -- and the artful combinations thereof -- largely set the bounds of possible cultural evolution (together, perhaps, with other cognitive features). Absent of more sophisticated forms of social learning, apes are unable to add novel traits, or to build cumulative traditions that progress beyond the boundary of `latent solutions'. Apes, but not humans, do not seem to copy -- or transmit -- traits beyond their ZLS (be it in the technical \citep{tennie2009ratcheting}, or social domain \citep{clay2017overimitation}). As said above, the appearance of cumulative culture can largely be accounted for by socially-facilitated reinnovation \citep{tennie2020zone}. That being said, there is some contrary evidence that suggests apes have limited capacities for some cumulative culture. Yet these capacities seem restricted to particular domains, with specific kinds of knowledge, and often in highly structured learning environments \citep{cladiere2014baboon,sasaki2017cumulative}. Absent these conditions, ape cultural evolution may be robustly bounded.

What might explain the transition between bounded ape culture and unbounded human culture? Though a full catalogue of important underlying processes has not yet been completed, a key capacity seems to be abilities for copying ``know-how'' -- that is, capacities for attending to, perhaps understanding, and copying/reconstructing the elements and interrelationships of \emph{any} particular behavior (including the making of artefacts; and of artefact structures themselves). Other relevant capacities - at least for modern humans - plausibly include language, and special types of teaching (especially those types of teaching that can transmit know-how).

ZLS research thus helps the current project in two ways. First, it helps to sharpen the notion of cultural evolutionary boundedness. Boundedness involves a limited exploration of cultural evolutionary space, due to minimal, lacking, or rarely expressed capacities for transmission, retention, or the production of modifications. Second, it helps to illuminate the devilish empirical issues involved in understanding the transition from boundedness to unboundedness. Focusing on the tall, wide, and unbounded cultural evolution of humans alone may not be helpful for understanding this transition \citep{buskellinpressmere}, but a combined focus that also includes understanding the patterns of change in evolutionary systems that ultimately fail to break away from boundedness may.

\subsection{Evolved Open-Endedness in Action}
According to \citeauthor{pattee2019evolved}, ``conditions for increased open-endedness must have been gradually acquired in the course of evolution" (\citeyear[p.~5]{pattee2019evolved}). In justifying this claim, \citeauthor{pattee2019evolved} point not only to concepts from the foundations of the modern synthesis \citep{haldane1932causes} and other more recent attempts to frame evolution as a progression of steps towards increased evolvability \citep{wagner1996perspective,wilson1997altruism,maynardsmith1995major,szathmary2015toward}, but also to numerous examples of evolved mechanisms that have ``significantly facilitated the open-endedness in the evolution of life" \citep[p.~6]{pattee2019evolved}. Notable amongst these examples are: 
\begin{itemize}
    \item the evolution of symbolic language spoken by humans, which are noted as being ``evolved from simpler, less open-ended languages" \citep[p.~6]{pattee2019evolved}.
    \item the formation of co-operative groups of increasing scale and complexity (colonies --> societies), with higher levels of organisational and institutional formation requiring the evolution of new mechanisms not previously seen in lower-level organisational entities.
    \item the evolution of new information-processing abilities, sensory modalities, and the brain, all providing organisms with new possibilities to explore and exploit.
\end{itemize}
From these examples it is clear that \citet{pattee2019evolved} consider what we describe as the evolution of culture (e.g. languages and social institutions) and the biological mechanism that support culture (e.g. the brain and culture supporting sensory modalities), as clear examples of evolved open-endedness. Therefore, we believe that in human cultural evolution (including ``dual-inheritance" and ``triple-inheritance") we have a real (and recent) example of evolved open-endedness in action. Below, we outline the case for human culture evolution as an instance of evolved open-endedness in action.  
Within cultural species more broadly, we can differentiate between different types of cultural evolution: bounded non-cumulative, bounded cumulative, unbounded non-cumulative, and unbounded cumulative. While cumulative culture may or may be uniquely human \citep{mesoudi2018cumulative}, unbounded cumulative culture plausibly is. Indeed, human cultural evolution appears to be the only instance of unbounded cumulative cultural evolution. 

Evidence suggests that the transition towards unbounded cumulative cultural evolution has taken place over the last few hundred thousand with the origin and evolution of \emph{Homo sapiens} \citep{stringer2016origin, stringer2017origin}, or even few million years with the advent on stone tool use in early \emph{Homo} \citep{lewis2016earlier}. We thus have, in both archaeological remains and in our genes, the record of this transition into open-ended cultural evolution. Exploring this transition is valuable, for it offers a compelling insight into the problems, solutions, processes and complex evolutionary dynamics that can jointly explain the emergence of a new open-ended evolutionary system. Though this is a particular instance, we suspect the concepts, tools, and ideas can be generalised.

This is not to say explaining the transition from primate ancestors to fully-fledged cultural hominins is easy. Anything but. Contemporary narratives point to a number of important changes that might have facilitated the evolution of a robust, quasi-independent system for cultural inheritance. These include changes in morphology (the bipedal stance, decreased gut size, increased crania), life history and population structure (social affiliation, intergenerational care, long developmental periods, extended family groups and social institutions), and cognitive attributes and machinery (greater executive control, social tolerance and attentiveness) \citep{sterelny2012evolved, sterelny2021pleistocene, klein2008career, kaplan2000theory, aiello1995expensive, grove2017environmental, anton2014evolution, ostrom1990governing, powers2013co,powers2016institutions}.

Just as important were cultural evolutionary feedback loops where early culture could facilitate selection for more and more effective social learning. Pre-modern hominin culture, for instance, generated an information environment seeded with cues as to how one should live. This includes ``scaffolded'' learning environments, where juveniles can learn in a relatively safe and low cost manner by interacting with the products of adult cooperation. These low-cost and safe learning environments could be increasingly supplemented with real-world experience, perhaps teaching, and experimentation as learners developed. Selection to improve capacities to navigate and explore this informational domain would in turn lead to greater informational structure in the world -— and thus to further selection. This general story is one of humans as ``evolved apprentices'' \citep{sterelny2012evolved}.

The story of how hominins escaped the ``boundedness'' of their primate relatives exploits this evolutionary feedback loop, increasing capacities for both tall and wide culture, and abilities to recognize ``task-independent'' properties of artefacts and behaviors that could be transferred and combined with other behaviors to generate new kinds of cultural traditions. These cognitive and cultural capacities could open up new evolutionary domains by exploiting novel affordances  \citep{arthur2009nature,derex2022exploit}. As a result, human technologies capture and put to use a collection of phenomena: for example, a car not only exploits the phenomenon that rolling objects produce much less friction than sliding ones (resulting in the use of wheels), but it also exploits the phenomenon that chemical substances (diesel, say) produce energy when burned \citep{arthur2009nature}. This discovery and exploitation of new solutions to old problems allows a potentially unbounded form of cumulative culture. We see evidence for the opening-up of new evolutionary search spaces, and the exploitation of new solutions in numerous domains within patent records \citep{bedau2019open, bedau2019patented, bedau2013minimal}.

Equally important is the way that human groups can support the increasing specialisation of skills and knowledge, the circulation of knowledge, and participation in collective endeavours -- pitching in on large or temporally distributed projects that could never be completed by a single agent in their own lifetime. These social features in turn could contribute to the changes in cognition, life history, and information dynamics discussed above. This is part of what some have called -- with various slight differences -- the cultural intelligence hypothesis \citep{herrmann2007specialized,vanschaik2011social,muthukrishna2018cultural}.

As this makes clear, the transition between a limited type of social learning and the more complex and open-ended form currently enjoyed by humans is a complex story. Despite this, complexity researchers in archaeology, comparative psychology, paleoanthropology, psychology, philosophers, and many others have been able to make progress on disentangling distinct causal pathways, and to show how these can be put together again to explain the evolution of a distinct system of open-ended evolution: human cultural evolution \citep{tomasello1999origins,boyd1985culture}. 

\section{Cultural Evolution, Open-Ended Evolution and Artificial Life}

Culture and cultural evolution have a long tradition in Artificial Life, appearing amongst both the grand challenges \citep{taylor1993artificial} and open problems \citep{bedau2000open} of the field, and spawning a regular workshop series at the Artificial Life conference \citep{marriott2018social}. It is therefore curious that open-ended cultural evolution has received relatively little attention as a possible avenue for fruitful research until recently (see \citet{bedau2019open}).

In the previous sections of this paper we have outlined many of the arguments and factors that we feel place cultural evolution firmly within the domain of open-ended evolution research. However, we also note a curious parallel between the work already taking place within the Artificial Life open-ended research community and the broader study of culture as an evolving system. A particular example of this can be seen in \citet{taylor2019evolutionary}, where three classes of novelty, all capable of generating open-ended evolution, are introduced: 1) \emph{exploratory novelty}, whereby existing traits are recombined to produce novel adaptations, 2) \emph{expansive novelty} resulting from the discovery and exploitation of new affordances, and 3) \emph{transformative novelty} resulting from the discovery of new state spaces, possibly via the exaptation of current traits. Within the cultural evolution literature we can see clear parallels with each of these classes: \emph{exploratory novelty} can be seen as a restricted process of cultural variation and accumulated modification within one domain or affordance (described as Type I cumulative cultural evolution by \citet{derex2022exploit}); \emph{expansive novelty} can be interpreted as an exploration of new affordances, expanding cultural evolution in to new domains (described as Type II cumulative cultural evolution by \citet{derex2022exploit}); and \emph{transformative novelty} can be viewed as movement into an n-dimensional state-space through the recombination and exaptation of existing cultural traits, enabling the creation and exploitation of new cultural and ecological niches. Examples of cultural exaptation abound in numerous domains, technology \citep{boyd2013cultural, bedau2019open, bedau2019patented} and pharmaceuticals \citep{andriani2015measuring} being two such examples.

It is evident that open-ended evolution research in artificial life and cultural evolution research have been speaking about very similar things; the types of novelty discussed by \citet{taylor2019evolutionary} and core aspects of cumulative cultural evolution outlined by \citet{derex2022exploit} and \citet{mesoudi2018cumulative} demonstrate such similarities. It should therefore be uncontroversial to suggest an open-ended evolutionary synthesis that combines genetic evolution, cultural evolution, and artificial evolution within a single theoretical framework. Combined with the exploratory work on open-ended technological innovation of \citet{bedau2019open, bedau2019patented}, the inclusion of social and cultural transitions emerging from earlier biological transitions within the major transitions framework \citep{maynardsmith1995major, szathmary2015toward, calcott2011major}, and the clear articulation of evidence for both biological and cultural mechanisms for the facilitation of evolved open-endedness \citep{pattee2019evolved}, we see a strong argument for the inclusion of cultural evolution within the broader framework of open-ended evolution.

In the sections below we argue that the transition from bounded to unbounded evolution, that is evident within the recent hominin evolutionary history, shines an important light on how evolved open-endedness might be achieved. We go on to consider tall and wide evolution within the context of the \citet{bedau1998classification} ``ALife Test'' and provide some initial thoughts on how this test could be further expanded to detect tall and wide adaptations in order to better delineate between the mechanisms driving (and halting) artificial evolutionary systems. Finally, we introduce a raft of new questions that the inclusion of cultural evolution under the framework of evolved open-endedness allows us to ask.

\subsection{Transitions from Bounded to Unbounded Evolution}
As we saw in section two, it is common to operationalize culture in informational terms: culture is information, embedded (or carried) by heterogeneous vehicles, that can be transmitted between agents \citep{richerson2005notbygenes}. On this understanding, one thread tying together the evolutionary history of hominin populations is an increase in and improvement of culturally transmitted information \citep{boyd1985culture}. This general observation has led some researchers to claim that culture represents a ``major transition'' in the sense of \citet{maynardsmith1995major} and \citet{szathmary2015toward}, building off the idea that such transitions involve changes in the quality and reliability of information transfer. For instance, \citet{waring2021longterm} argue that human cultural groups are a new kind of evolutionary individual, suggesting that cultural selection pressures now vastly outweigh biological selection pressures in determining the course of human diversification and change.

Waring and Wood’s arguments interpret the major transitions framework in a particularly strong way. This takes transitions to involve the stabilization of a new evolutionary individual, here, a cultural group \citep{mcshea2011miscellaneous}. But one need not understand the framework in this ``unified'' way \citep{michod1999dynamics}. Instead, transitions may involve modifications of the ``core elements of the evolutionary process itself'' \citep[p.~4]{calcott2011major}, irrespective of introducing a new level or kind of selection process \citep{godfreysmith2009darwinian}. Thus, even if one is sceptical about cultural group selection (see, for instance, \citet{chellappoo2022selection}) one can usefully understand the introduction and refinement of cultural evolution using the ideas and machinery of the major transition literature \citep{maynardsmith1995major,szathmary2015toward, calcott2011major}.

We conceive ``open-endedness'' through this more expansive understanding. It characterises an increase of informational content that can be (or is) transmitted in a given domain, potentially reflecting coordinated or piecemeal changes to the rate, increased quantity, or kind of variation that can be generated. In so doing, we follow \citet{pattee2019evolved}:
``[o]ver time both biological adaptations that enable more complex and open-ended social and cultural behaviors (bigger brains, opposable thumbs, changes in the shape of the larynx, ...), and cultural adaptations that open up access to new domains of knowledge (symbolic language, the scientific method, music and art, complex social institutions, ...) have been selected for in a clear demonstration of selection in favour of open-endedness, with this same selection pressure being seemingly absent in our closest genetic relatives''.

\subsection{Cultural Evolution and the ``ALife Test'' for Open-Endedness}
Determining whether an evolutionary system exhibits unbounded evolutionary dynamics is still arguably the primary concern of open-ended evolution research. Without the ability to judge whether a system is open-ended, how can open-endedness be understood to any useful degree? Despite a general lack of use, we are of the opinion that the classification system of long-term evolutionary dynamics devised by \citet{bedau1998classification} (sometimes known as the ``ALife Test'' for open-endedness) provides us with the best method for determining whether an evolutionary system exhibits unbounded evolutionary dynamics. However, we believe some of the key features of cultural evolution -- wide vs. tall evolution, transition from bounded to unbounded evolution, and evolved open-endedness -- may necessitate some refinement of the ``ALife Test''.

The three primary measures of evolutionary activity described in \citet{bedau1998classification} are 1) the diversity of traits within the system at any given time, 2) the amount of ``new evolutionary activity'' observed in the system over time (i.e., the creation and maintenance of new adaptive traits), and 3) the mean cumulative activity of traits (i.e., the number of traits observed to date divided by the current diversity of traits in the system). For a system to exhibit unbounded evolutionary dynamics it would need to always demonstrate positive new evolutionary activity (i.e. new traits are being created and maintained), alongside either unbounded diversity (as time progresses the number of traits maintained in the system continues to grow) and/or unbounded mean cumulative activity. 

What these measures of evolutionary activity do not take into account is whether the new activity is a result of cumulative evolutionary processes, non-cumulative evolutionary processes, or recombinative processes. These distinctions matter because they can begin to shed light on \emph{how} a system has progressed toward, and ultimately achieved, open-endedness. For instance, would we expect to see a ``building-out'' of wide adaptations (as seems to be the case in hominin cultural evolution) before the emergence of tall accumulated modifications, ultimately leading to the combination of traits from disparate evolutionary lineages forming recombinative adaptations (wide evolution providing the raw material for exploratory and expansive evolution as per \citet{taylor2019evolutionary}? Or are there numerous different pathways to open-endedness which can only be understood by breaking down the nature of the evolutionary patterns of change, adaptive processes, substrate and mechanisms underpinning these evolutionary systems?

\subsection{New Questions in Open-Endedness}
Once we consider the implications and nature of cultural evolution from an open-ended evolution perspective we can begin to ask new and important questions about evolved open-endedness, human cultural evolution, and the underpinning dynamics of all evolutionary systems. These questions include, but are not limited to:
\begin{itemize}
    \item Do the mechanisms underpinning cultural evolution more easily lead to open-endedness than those underpinning genetic evolution? Or vice-versa?
    \item What happens when a bounded aspect of an evolutionary system (e.g. animal cultural evolution) comes up against an unbounded aspect of the same evolutionary system (e.g. human open-ended cultural evolution)? Is there a sudden pressure for evolved open-endedness to emerge amongst species that have so far only exhibited bounded cultural evolution? And does the emergence of open-endedness always lead to the extinction of its bounded counterpart?
    \item Are there any bounded aspects of human cultural evolution? And could there also be bounded aspects of genetic evolution?
    \item Does an evolutionary system need to be cumulative to be open-ended, or is it possible to have non-cumulative open-ended evolution? Note: If major transitions are one of the primary behavioral hallmarks of an open-ended evolutionary system \citep{taylor2016open}, and major transitions build up incrementally from one another (each transition is dependent on subsequent levels), this would imply that open-ended evolution must result from a cumulative evolutionary process. But is it possible to generate open-ended evolution without cumulative major transitions and could major transitions be the result of numerous independent innovations?
    \item Are cumulative evolutionary systems always open-ended? The numerous cases outlined in \citet{mesoudi2018cumulative} would suggest not, nor do the criteria for cumulative cultural evolution necessitate an open-ended system (or logically lead to the conclusion that open-ended evolution is an unavoidable end point).
    \item What features of cultural evolution are common to all evolutionary systems capable of generating the open-ended evolution of novelty?
    \item Is an open-ended evolutionary synthesis which accommodates cultural evolution alongside genetic evolution and artificial evolution viable and/or desirable?
    \item Is niche construction necessary for open-ended evolution? And are the autocatalytic processes resulting from the interplay between numerous interdependent evolutionary systems (see ``triple inheritance" \citep{laland2000triple}) necessary for open-endedness?
\end{itemize}

\section{Conclusion}
In this paper we set out to outline culture as an evolutionary system and argue for its inclusion within the broader framework of evolved open-endedness. In order to make these arguments we provided numerous examples of the unique aspects of cultural evolution that make it a fascinating counterpoint to biological evolution, but we also maintain a direct link between the core algorithmic features of biological evolution and cultural evolution. We went on to discuss the key features and dynamics of cultural evolution, including: tall, wide, cumulative and non-cumulative evolution, transitions from bounded to unbounded evolution, dual and triple inheritance, evolved open-endedness, major transitions, and the ZLS theory. Each of these features provide new insights into the nature of another model evolutionary system.

Going forward we believe two lines of enquiry are necessary to fully develop cultural evolution as an integral part of open-ended evolution research. 1) Following on from the work of \citet{bedau2019open}, we believe an application of the ``ALife Test'' to the vast number of available cultural evolution datasets, across numerous species, would be informative for both the open-evolution community and the cultural evolution community. 2) Including mechanisms of cultural transmission and the unique features of cultural evolution within artificial evolutionary models aimed at addressing the question of open-endedness -- this may involve the modelling of culture as an independent system, or the inclusion of culture alongside genetic (and environmental) inheritance. To enable these two lines of enquiry we believe some work on the refinement of the ``ALife Test'' is necessary, as is the development of tall- wide-recombinative evolutionary theory, and more interdisciplinary dialogue between the fields of Cultural Evolution and Artificial Life.

\section{Acknowledgements}
We would like to thank the members of the Cultural Evolution Online (CEO) Discord group who have provided advice and insights that have helped us form the views contained in this paper, and James Winters and Mathieu Charbonneau for their involvement in regular discussions with the authors on the topics addressed in this paper. We would also like to thank our two reviewers; both provided a series of constructive comments that have unquestionably improved this paper. Finally we would like to thank the organisers of the OEE4 workshop and the guest editors of this special issue for providing us with the opportunity to present this work.

\printbibliography

@inproceedings{andriani2015measuring,
  title={Measuring exaptation in the pharmaceutical industry},
  author={Andriani, Pierpaolo and Ali, Ayfer H and Mastrogiorgio, Mariano},
  booktitle={Academy of Management Proceedings},
  volume={2015},
  number={1},
  pages={17085},
  year={2015},
  organization={Academy of Management Briarcliff Manor, NY 10510},
  doi = {10.5465/ambpp.2015.17085abstract}
}

@article{aiello1995expensive,
  title={The expensive-tissue hypothesis: the brain and the digestive system in human and primate evolution},
  author={Aiello, Leslie C and Wheeler, Peter},
  journal={Current anthropology},
  volume={36},
  number={2},
  pages={199--221},
  year={1995},
  publisher={University of Chicago Press},
  doi = {10.1086/204350}
}

@article{anton2014evolution,
  title = {Evolution of early Homo: An integrated biological perspective},
  author = {Ant{\'o}n, Susan C and Potts, Richard and Aiello, Leslie C},
  journal = {Science},
  volume = {345},
  number = {6192},
  pages = {1236828},
  year = {2014},
  publisher = {American Association for the Advancement of Science},
  doi = {10.1126/science.1236828}
}

@article{aplin2019birds,
    title = {Culture and cultural evolution in birds: a review of the evidence},
    author={Aplin, Lucy M},
    journal = {Animal Behaviour},
    volume = {147},
    pages = {179-187},
    year = {2019},
    issn = {0003-3472},
    doi = {https://doi.org/10.1016/j.anbehav.2018.05.001}
}

@book{arthur2009nature,
  title={The nature of technology: What it is and how it evolves},
  author={Arthur, W Brian},
  year={2009},
  publisher={Simon and Schuster}
}

@article{baehren2022saying,
  title={Saying “goodbye” to the conundrum of leave-taking: a cross-disciplinary review},
  author={Baehren, Lucy},
  journal={Humanities and Social Sciences Communications},
  volume={9},
  number={1},
  pages={1--13},
  year={2022},
  publisher={Palgrave},
  doi={10.1057/s41599-022-01061-3}
}

@article{bandini2017scooping,
    title = {Spontaneous reoccurrence of “scooping”, a wild tool-use behaviour, in na\"{i}ve chimpanzees},
    author = {Elisa Bandini and Claudio Tennie},
    year = {2017},
    journal = {PeerJ},
    volume = {5},
    number = {e3814},
    doi = {10.7717/peerj.3814}
}

@article{banea1992shortcuts,
  title={Shortcuts in cassava processing and risk of dietary cyanide exposure in {Z}aire},
  author={Banea, Mayambu and Poulter, Nigel H and Rosling, Hans},
  journal={Food and Nutrition Bulletin},
  volume={14},
  number={2},
  pages={1--7},
  year={1992},
  publisher={SAGE Publications Sage CA: Los Angeles, CA},
  doi={10.1177/156482659201400201}
}

@inproceedings{bedau1998classification,
  title={A classification of long-term evolutionary dynamics},
  author={Bedau, Mark A and Snyder, Emile and Packard, Norman H},
  booktitle={Artificial Life VI: Proceedings of the Sixth International Conference on Artificial Life},
  pages={228--237},
  year={1998},
  editor={Christoph Adami and Richard K. Belew and Hiroaki Kitano and Charles E. Taylor},
  publisher={MIT Press}
}

@article{bedau2000open,
  title={Open problems in artificial life},
  author={Bedau, Mark A and McCaskill, John S and Packard, Norman H and Rasmussen, Steen and Adami, Chris and Green, David G and Ikegami, Takashi and Kaneko, Kunihiko and Ray, Thomas S},
  journal={Artificial life},
  volume={6},
  number={4},
  pages={363--376},
  year={2000},
  publisher={MIT Press},
  doi={10.1162/106454600300103683}
}

@article{bedau2013minimal,
  title={Minimal memetics and the evolution of patented technology},
  author={Bedau, Mark A},
  journal={Foundations of science},
  volume={18},
  number={4},
  pages={791--807},
  year={2013},
  publisher={Springer}
}

@incollection{bedau2019patented,
  title={Patented technology as a model system for cultural evolution},
  author={Bedau, Mark A},
  booktitle = {Beyond the Meme: Development and Structure in Cultural Evolution},
  editor = {Alan Love and William Wimsatt},
  series = {Minnesota Studies in the Philosophy of Science},
  volume ={22},
  pages = {237-260},
  year={2019},
  publisher={U of Minnesota Press}
}

@article{bedau2019open,
    title = {Open-Ended Technological Innovation},
    author = {Mark A. Bedau and Nicholas Gigliotti and Tobias Janssen and Alec Kosik and Ananthan Nambiar and Norman Packard},
    year = {2019},
    journal = {Artificial Life},
    volume = {25},
    number = {1},
    pages = {33-49},
    doi = {10.1162/artl_a_00279}
}

@article{boesch1990tool,
  title={Tool use and tool making in wild chimpanzees},
  author={Boesch, Christophe and Boesch, Hedwige},
  journal={Folia primatologica},
  volume={54},
  number={1-2},
  pages={86--99},
  year={1990},
  publisher={Karger Publishers},
  doi={10.1159/000156428}
}

@inproceedings{borg2021evolved,
  title={Evolved Open-Endedness in Cultural Evolution},
  author={Borg, James and Powers, Simon T},
  booktitle={The Fourth Workshop on Open-Ended Evolution},
  year={2021}
}

@book{boyd1985culture,
    title = {Culture and the evolutionary process},
    author = {Robert Boyd and Peter J. Richerson},
    year = {1985},
    publisher = {University of Chicago press}
}

@incollection{boyd2013cultural,
  title={The cultural evolution of technology},
  author={Boyd, Robert and Richerson, Peter J and Henrich, Joseph},
  booktitle={Cultural evolution: society, technology, language, and religion},
  editor={Peter J. Richerson and Morton H. Christiansen},
  volume={12},
  pages={119--142},
  year={2013},
  publisher={MIT Press Cambridge, MA},
  doi={10.7551/mitpress/9780262019750.003.0007}
}

@article{bradbury2011mild,
  title={Mild methods of processing cassava leaves to remove cyanogens and conserve key nutrients},
  author={Bradbury, J Howard and Denton, Ian C},
  journal={Food Chemistry},
  volume={127},
  number={4},
  pages={1755--1759},
  year={2011},
  publisher={Elsevier},
  doi={10.1016/j.foodchem.2011.02.053}
}

@inproceedings{bullinaria2010memes,
  title={Memes in Artificial Life Simulations of Life History Evolution},
  author={Bullinaria, John A},
  booktitle={Proceedings of the ALife XII Conference},
  pages={823--830},
  year={2010}
}

@article{bull2000meme,
  title={On meme--gene coevolution},
  author={Bull, Larry and Holland, Owen and Blackmore, Susan},
  journal={Artificial life},
  volume={6},
  number={3},
  pages={227--235},
  year={2000},
  publisher={MIT Press},
  doi = {10.1162/106454600568852}
}

@article{buskell2018religion,
    title={Causes of Cultural Disparity: Switches, Tuners, and the Cognitive Science of Religion},
    author={Buskell, Andrew},
    journal={Philosophical Psychology},
    volume={31},
    number={8},
    pages={1239--1264},
    year={2018},
    doi={10.1080/09515089.2018.1485888}
}

@article{buskell2019systems,
  title={A systems approach to cultural evolution},
  author={Buskell, Andrew and Enquist, Magnus and Jansson, Fredrik},
  journal={Palgrave Communications},
  volume={5},
  number={1},
  pages={1--15},
  year={2019},
  publisher={Palgrave},
  doi={10.1057/s41599-019-0343-5}
}

@article{buskellinpressmere,
    title={Mere recurrence and cumulative culture at the margins},
    author={Andrew Buskell and Claudio Tennie},
    journal={The British Journal for the Philosophy of Science},
    year={in press}
}

@article{buskellForthcomplex,
    title={Cumulative Culture and Complex Cultural Traditions},
    author={Buskell, Andrew},
    year={forthcoming},
    journal={Mind \& Language},
    doi={10.1111/mila.12335}
}

@article{calcott2009lineage,
    title={Lineage Explanations: Explaining How Biological Mechanisms Change},
    author={Calcott, Brett},
    journal={The British Journal for the Philosophy of Science},
    volume={60},
    number={1},
    pages={51--78},
    year={2009},
    doi={10.1093/bjps/axn047}
}

@book{calcott2011major,
  title={The major transitions in evolution revisited},
  author={Calcott, Brett and Sterelny, Kim},
  year={2011},
  publisher={MIT Press}
}

@article{cardoso2005processing,
  title={Processing of cassava roots to remove cyanogens},
  author={Cardoso, A Paula and Mirione, Estevao and Ernesto, Mario and Massaza, Fernando and Cliff, Julie and Haque, M Rezaul and Bradbury, J Howard},
  journal={Journal of Food Composition and Analysis},
  volume={18},
  number={5},
  pages={451--460},
  year={2005},
  publisher={Elsevier},
  doi={10.1016/j.jfca.2004.04.002}
}

@book{cavalli1981cultural,
    title={Cultural transmission and evolution: A quantitative approach},
    author={Luigi Luca Cavalli-Sforza and Marcus W. Feldman},
    year={1981},
    publisher={Princeton University Press}
}

@misc{CES,
    author = {{Cultural Evolution Society}},
    title   = {{What is Cultural Evolution}?},
    year    =  {2021},
    url =   {https://culturalevolutionsociety.org/story/What_is_Cultural_Evolution},
    urldate = {2021-06-06},
    Howpublished = {\url{https://culturalevolutionsociety.org/story/What_is_Cultural_Evolution}},
    note = {Last checked on June~06, 2021}
}

@inproceedings{channon2001passing,
  title={Passing the ALife test: Activity statistics classify evolution in Geb as unbounded},
  author={Channon, Alastair},
  booktitle={Lecture Notes in Computer Science vol. 2159: Advances in Artificial Life, {ECAL} 2001 },
  pages={417--426},
  year={2001},
  editor={Jozef Kelemen, Petr Sosik},
  organization={Springer},
  doi={10.1007/3-540-44811-X_45}
}

@inproceedings{channon2003improving,
  title={Improving and still passing the ALife test: Component-normalised activity statistics classify evolution in Geb as unbounded},
  author={Channon, Alastair},
  booktitle={Artificial Life {VIII}: Proceedings of the Eighth International Conference on Artificial Life},
  pages={173--181},
  year={2003},
  editor={Russell K. Standish and Mark A. Bedau and Hussein A. Abbass},
  publisher={MIT Press}
}

@article{channon2006unbounded,
  title={Unbounded evolutionary dynamics in a system of agents that actively process and transform their environment},
  author={Channon, Alastair},
  journal={Genetic Programming and Evolvable Machines},
  volume={7},
  number={3},
  pages={253--281},
  year={2006},
  publisher={Springer},
  doi={10.1007/s10710-006-9009-3}
}

@article{charbonneau2016modularity,
  title={Modularity and recombination in technological evolution},
  author={Charbonneau, Mathieu},
  journal={Philosophy \& Technology},
  volume={29},
  number={4},
  pages={373--392},
  year={2016},
  publisher={Springer},
  doi={10.1007/s13347-016-0228-0}
}

@article{chellappoo2022selection,
    title = {When can cultural selection explain adaptation?},
    author = {Azita Chellappoo},
    journal = {Biology \& Philosophy},
    year = {2022},
    volume = {37},
    doi = {10.1007/s10539-021-09831-0}
}

@article{cladiere2014baboon,
    author = {Nicolas Claidi\`{e}re and Kenny Smith and Simon Kirby and Jo\"{e}l Fagot},
    title = {Cultural evolution of systematically structured behaviour in a non-human primate},
    journal = {Proceedings of the Royal Society B: Biological Sciences},
    volume = {281},
    number = {1797},
    pages = {20141541},
    year = {2014},
    doi = {10.1098/rspb.2014.1541}
}

@book{dawkins1976selfish,
  title={The selfish gene},
  author={Dawkins, Richard},
  year={1976},
  publisher={Oxford University Press}
}

@article{dean2014human,
  title={Human cumulative culture: a comparative perspective},
  author={Dean, Lewis G and Vale, Gill L and Laland, Kevin N and Flynn, Emma and Kendal, Rachel L},
  journal={Biological reviews},
  volume={89},
  number={2},
  pages={284--301},
  year={2014},
  publisher={Wiley Online Library},
  doi={10.1111/brv.12053}
}

@book{dennett1996darwin,
  title={Darwin’s Dangerous Idea: Evolution and the Meanings of Life},
  author={Dennett, Daniel C},
  year={1996},
  publisher={Penguin}
}

@article{derex2022exploit,
    author = {Maxime Derex},
    title = {Human cumulative culture and the exploitation of natural phenomena},
    journal = {Philosophical Transactions of the Royal Society B: Biological Sciences},
    volume = {377},
    number = {1843},
    pages = {20200311},
    year = {2022},
    doi = {10.1098/rstb.2020.0311}
}

@article{duranti1997universal,
  title={Universal and culture-specific properties of greetings},
  author={Duranti, Alessandro},
  journal={Journal of linguistic Anthropology},
  volume={7},
  number={1},
  pages={63--97},
  year={1997},
  publisher={Wiley Online Library},
  doi={10.1525/jlin.1997.7.1.63}
}

@article{enquist2010one,
  title={One cultural parent makes no culture},
  author={Enquist, Magnus and Strimling, Pontus and Eriksson, Kimmo and Laland, Kevin and Sjostrand, Jonas},
  journal={Animal Behaviour},
  volume={79},
  number={6},
  pages={1353--1362},
  year={2010},
  publisher={Elsevier},
  doi={10.1016/j.anbehav.2010.03.009}
}

@article{fitch2011unity,
  title={Unity and diversity in human language},
  author={Fitch, W Tecumseh},
  journal={Philosophical Transactions of the Royal Society B: Biological Sciences},
  volume={366},
  number={1563},
  pages={376--388},
  year={2011},
  publisher={The Royal Society},
  doi={10.1098/rstb.2010.0223}
}

@article{gabora2017autocatalytic,
  title={Autocatalytic networks in cognition and the origin of culture},
  author={Gabora, Liane and Steel, Mike},
  journal={Journal of Theoretical Biology},
  volume={431},
  pages={87--95},
  year={2017},
  publisher={Elsevier},
  doi={10.1016/j.jtbi.2017.07.022}
}

@article{griffin2004social,
  title={Social learning about predators: a review and prospectus},
  author={Griffin, Andrea S},
  journal={Animal Learning \& Behavior},
  volume={32},
  number={1},
  pages={131--140},
  year={2004},
  publisher={Springer},
  doi={10.3758/BF03196014}
}

@article{grove2011speciation,
  title={Speciation, diversity, and Mode 1 technologies: The impact of variability selection},
  author={Grove, Matt},
  journal={Journal of Human Evolution},
  volume={61},
  number={3},
  pages={306--319},
  year={2011},
  publisher={Elsevier},
  doi={10.1016/j.jhevol.2011.04.005}
}

@article{grove2017environmental,
  title={Environmental complexity, life history, and encephalisation in human evolution},
  author={Grove, Matt},
  journal={Biology \& Philosophy},
  volume={32},
  number={3},
  pages={395--420},
  year={2017},
  publisher={Springer},
  doi={10.1007/s10539-017-9564-4}
}

@incollection{henrich2007dual,
    title={Dual inheritance theory: The evolution of human cultural capacities and cultural evolution},
    author={Henrich, Joseph and McElreath, Richard},
    booktitle = {Oxford handbook of evolutionary psychology},
    editor = {Louise Barrett and Robin Dunbar},
    pages = {555--570},
    publisher = {Oxford University Press},
    year = {2007}
}

@article{henrich2010evolution,
  title={The evolution of cultural adaptations: {F}ijian food taboos protect against dangerous marine toxins},
  author={Henrich, Joseph and Henrich, Natalie},
  journal={Proceedings of the Royal Society B: Biological Sciences},
  volume={277},
  number={1701},
  pages={3715--3724},
  year={2010},
  publisher={The Royal Society},
  doi={10.1098/rspb.2010.1191}
}

@book{henrich2015secret,
    title = {The secret of our success. How culture is driving human evolution, domesticating our species, and making us smarter},
    author = {Joseph Henrich},
    year = {2015},
    publisher = {Princeton University Press},
    doi={10.1515/9781400873296}
}

@book{heyes2018gadgets,
    title={Cognitive Gadgets: The Cultural Evolution of Thinking},
    author={Heyes, Cecilia},
    year={2018},
    publisher={Harvard University Press}
}

@article{howard2017frequency,
  title={Frequency-dependent female genital cutting behaviour confers evolutionary fitness benefits},
  author={Howard, Janet A and Gibson, Mhairi A},
  journal={Nature Ecology \& Evolution},
  volume={1},
  number={3},
  pages={1--6},
  year={2017},
  publisher={Nature Publishing Group},
  doi={10.1038/s41559-016-0049}
}

@article{janik2000different,
  title={The different roles of social learning in vocal communication},
  author={Janik, Vincent M and Slater, Peter JB},
  journal={Animal behaviour},
  volume={60},
  number={1},
  pages={1--11},
  year={2000},
  publisher={Elsevier},
  doi={10.1006/anbe.2000.1410}
}

@book{godfreysmith2009darwinian,
    title = {Darwinian Populations and Natural Selection},
    author = {Peter Godfrey-Smith},
    year = {2009},
    publisher = {Oxford University Press}
}

@article{kaplan2000theory,
    title={A Theory of Human Life History Evolution},
    author={Kaplan, H and Hill, K and Lancaster, J and Hurtado, A M},
    journal={Evolutionary Anthropology},
    volume={9},
    pages={156--185},
    year={2000},
    doi={10.1002/1520-6505(2000)9:4<156::AID-EVAN5>3.0.CO;2-7}
}

@book{klein2008career,
    title={The Human Career},
    author={Klein, R G},
    edition={3rd},
    publisher={University of Chicago Press},
    year={2008}
}

@article{kolodny2015evolution,
  title={Evolution in leaps: the punctuated accumulation and loss of cultural innovations},
  author={Kolodny, Oren and Creanza, Nicole and Feldman, Marcus W},
  journal={Proceedings of the National Academy of Sciences},
  volume={112},
  number={49},
  pages={E6762--E6769},
  year={2015},
  publisher={National Acad Sciences},
  doi={10.1073/pnas.1520492112}
}

@article{laland2000triple, 
    title={Niche construction, biological evolution, and cultural change}, 
    doi ={10.1017/S0140525X00002417}, 
    journal={Behavioral and Brain Sciences}, 
    publisher={Cambridge University Press}, 
    author={Kevin N Laland and John Odling-Smee and Marcus W Feldman},
    year={2000}, 
    volume = {75},
    pages={131–175}
}

@article{laland2011cultural,
  title={Cultural niche construction: An introduction},
  author={Laland, Kevin N and O’Brien, Michael J},
  journal={Biological Theory},
  volume={6},
  number={3},
  pages={191--202},
  year={2011},
  publisher={Springer},
  doi={10.1007/s13752-012-0026-6}
}

@inproceedings{langton1989artificial,
    title={Artificial Life},
    author={Langton, Christopher G},
    booktitle={Artificial Life: proceedings of an interdisciplinary workshop on the Synthesis and Simulation of Living Systems, Los Alamos, 1987},
    year={1989},
    pages={1--47},
    editor={Langton, Christopher G},
    publisher={Addison-Wesley}
}

@article{lehman2020surprising,
  title={The surprising creativity of digital evolution: A collection of anecdotes from the evolutionary computation and artificial life research communities},
  author={Joel Lehman and Jeff Clune and Dusan Misevic and Christoph Adami and Lee Altenberg and Julie Beaulieu and Peter J. Bentley and Samuel Bernard and Guillaume Beslon and David M. Bryson and Nick Cheney and Patryk Chrabaszcz and Antoine Cully and Stephane Doncieux and Fred C. Dyer and Kai Olav Ellefsen and Robert Feldt and Stephan Fischer and Stephanie Forrest and Antoine Fŕenoy and Christian Gagńe and Leni Le Goff and Laura M. Grabowski and Babak Hodjat and Frank Hutter and Laurent Keller and Carole Knibbe and Peter Krcah and Richard E. Lenski and Hod Lipson and Robert MacCurdy and Carlos Maestre and Risto Miikkulainen and Sara Mitri and David E. Moriarty and Jean-Baptiste Mouret and Anh Nguyen and Charles Ofria and Marc Parizeau and David Parsons and Robert T. Pennock and William F. Punch and Thomas S. Ray and Marc Schoenauer and Eric Schulte and Karl Sims and Kenneth O. Stanley and François Taddei and Danesh Tarapore and Simon Thibault and Richard Watson and Westley Weimer and Jason Yosinski},
  journal={Artificial life},
  volume={26},
  number={2},
  pages={274--306},
  year={2020},
  publisher={MIT Press},
  doi={10.1162/artl_a_00319
}
}

@article{macdonald2021middle,
  title={Middle Pleistocene fire use: The first signal of widespread cultural diffusion in human evolution},
  author={MacDonald, Katharine and Scherjon, Fulco and van Veen, Eva and Vaesen, Krist and Roebroeks, Wil},
  journal={Proceedings of the National Academy of Sciences},
  volume={118},
  number={31},
  year={2021},
  publisher={National Acad Sciences},
  doi={10.1073/pnas.2101108118}
}

@article{marriott2018social,
  title={Social learning and cultural evolution in artificial life},
  author={Marriott, Chris and Borg, James M and Andras, Peter and Smaldino, Paul E},
  journal={Artificial Life},
  volume={24},
  number={1},
  pages={5--9},
  year={2018},
  publisher={MIT Press},
  doi={10.1162/ARTL_a_00250}
}

@book{maynardsmith1995major,
    title = {The Major Transitions in Evolution},
    author = {John {Maynard Smith} and E\"{o}rs Szathm\'{a}ry},
    year = {1995},
    publisher = {Oxford University Press}
}

@article{mckerracher2016food,
  title={Food aversions and cravings during pregnancy on {Y}asawa {I}sland, {F}iji},
  author={McKerracher, Luseadra and Collard, Mark and Henrich, Joseph},
  journal={Human Nature},
  volume={27},
  number={3},
  pages={296--315},
  year={2016},
  publisher={Springer},
  doi={10.1007/s12110-016-9262-y}
}

@incollection{mcshea2011miscellaneous,
  title={The miscellaneous transitions in evolution},
  author={McShea, Daniel W and Simpson, Carl},
  booktitle={The major transitions in evolution revisited},
  pages={19--34},
  year={2011},
  editor={Brett Calcott and Kim Sterelny},
  publisher={MIT Press Cambridge, Massachusetts}
}

@article{mesoudi2006unified,
    title = {Towards a unified science of cultural evolution},
    author = {Alex Mesoudi and Andrew Whiten and Kevin N Laland},
    journal = {Behavioral and Brain Sciences},
    volume={29},
    pages={329-383},
    year={2006},
    doi = {10.1017/S0140525X06009083}
}

@book{mesoudi2011cultevol,
    title = {Cultural Evolution: How {D}arwinian Theory can Explain Human Culture \& Synthesize the Social Sciences},
    author = {Alex Mesoudi},
    year = {2011},
    publisher = {University of Chicago Press}
}

@article{mesoudi2018cumulative,
    title = {What is cumulative cultural evolution?},
    author = {Alex Mesoudi and Alex Thornton},
    journal = {Proceedings of the Royal Society B},
    volume = {285},
    number = {1880},
    pages = {20180712},
    year = {2018},
    doi = {10.1098/rspb.2018.0712}
}

@book{michod1999dynamics,
    title = {Darwinian Dynamics},
    author = {Richard Michod},
    year = {1999},
    publisher = {Princeton University Press}
}

@article{muthukrishna2018cultural,
  title={The Cultural Brain Hypothesis: How culture drives brain expansion, sociality, and life history},
  author={Muthukrishna, Michael and Doebeli, Michael and Chudek, Maciej and Henrich, Joseph},
  journal={PLoS computational biology},
  volume={14},
  number={11},
  pages={e1006504},
  year={2018},
  publisher={Public Library of Science San Francisco, CA USA},
  doi={10.1371/journal.pcbi.1006504}
}

@article{neadle2017food,
  title={Food cleaning in gorillas: social learning is a possibility but not a necessity},
  author={Neadle, Damien and Allritz, Matthias and Tennie, Claudio},
  journal={PLoS One},
  volume={12},
  number={12},
  pages={e0188866},
  year={2017},
  publisher={Public Library of Science San Francisco, CA USA},
  doi={10.1371/journal.pone.0188866}
}

@book{ostrom1990governing,
  title={Governing the commons: The evolution of institutions for collective action},
  author={Ostrom, Elinor},
  year={1990},
  publisher={Cambridge university press}
}

@article{pattee2019evolved,
    title = {Evolved Open-Endedness, Not Open-Ended Evolution},
    author = {Howard H. Pattee and Hiroki Sayama},
    year = {2019},
    journal = {Artificial Life},
    volume = {25},
    number = {1},
    pages = {4--8},
    doi = {10.1162/artl_a_00276}
}

@article{perry2011traditions,
    title = {Social traditions and social learning in capuchin monkeys (\emph{Cebus})},
    author = {Susan Perry},
    journal = {Philosophical Transactions of the Royal Society B: Biological Sciences},
    volume = {366},
    number = {1567},
    pages = {988--996},
    year = {2011},
    doi = {10.1098/rstb.2010.0317}
}

@article{potts2013hominin,
  title={Hominin evolution in settings of strong environmental variability},
  author={Potts, Richard},
  journal={Quaternary Science Reviews},
  volume={73},
  pages={1--13},
  year={2013},
  publisher={Elsevier},
  doi={10.1016/j.quascirev.2013.04.003}
}

@article{powers2013co,
  title={The co-evolution of social institutions, demography, and large-scale human cooperation},
  author={Powers, Simon T and Lehmann, Laurent},
  journal={Ecology letters},
  volume={16},
  number={11},
  pages={1356--1364},
  year={2013},
  publisher={Wiley Online Library},
  doi={10.1111/ele.12178}
}

@article{powers2016institutions,
  title={How institutions shaped the last major evolutionary transition to large-scale human societies},
  author={Powers, Simon T and {van Schaik}, Carel P and Lehmann, Laurent},
  journal={Philosophical Transactions of the Royal Society B: Biological Sciences},
  volume={371},
  number={1687},
  year={2016},
  publisher={The Royal Society},
  doi={10.1098/rstb.2015.0098}
}

@book{richerson2005notbygenes,
    title = {Not By Genes Alone},
    author = {Richerson, Peter J and Boyd, Robert},
    year = {2005},
    publisher = {University of Chicago Press}
}

@article{richerson2009tribal,
  title={Tribal social instincts and the cultural evolution of institutions to solve collective action problems},
  author={Richerson, Peter J and Henrich, Joe},
  journal={Context and the Evolution of Mechanisms for Solving Collective Action Problems Paper},
  year={2009},
  doi={10.2139/ssrn.1368756}
}

@article{rosenberg2017social,
  title={Why social science is biological science},
  author={Rosenberg, Alex},
  journal={Journal for General Philosophy of Science},
  volume={48},
  number={3},
  pages={341--369},
  year={2017},
  publisher={Springer},
  doi={10.1007/s10838-017-9365-0}
}

@article{sanz2010chimpanzees,
  title={Chimpanzees prey on army ants with specialized tool set},
  author={Sanz, Crickette M and Sch{\"o}ning, Caspar and Morgan, David B},
  journal={American Journal of Primatology: Official Journal of the American Society of Primatologists},
  volume={72},
  number={1},
  pages={17--24},
  year={2010},
  publisher={Wiley Online Library},
  doi={10.1002/ajp.20744}
}

@article{sasaki2017cumulative,
  title = {Cumulative culture can emerge from collective intelligence in animal groups},
  author = {Sasaki, Takao and Biro, Dora},
  journal = {Nature communications},
  volume = {8},
  number = {1},
  pages = {1--6},
  year = {2017},
  publisher = {Nature Publishing Group},
  doi = {10.1038/ncomms15049}
}

@article{sosis2004adaptive,
  title={The adaptive value of religious ritual: Rituals promote group cohesion by requiring members to engage in behavior that is too costly to fake},
  author={Sosis, Richard},
  journal={American scientist},
  volume={92},
  number={2},
  pages={166--172},
  year={2004},
  publisher={JSTOR},
  doi={10.1511/2004.46.928}
}

@book{sterelny2012evolved,
  title={The evolved apprentice},
  author={Sterelny, Kim},
  year={2012},
  publisher={MIT press}
}

@book{sterelny2021pleistocene,
    title={The Pleistocene Social Contract},
    author={Sterelny, Kim},
    year={2021},
    publisher={Oxford University Press}
}

@article{szathmary2015toward,
    title = {Toward major evolutionary transitions theory 2.0},
    author = {E\"{o}rs Szathm\'{a}ry},
    year = {2015},
    journal = {Proceedings of the National Academy of Sciences},
    volume = {112},
    number = {33},
    pages = {10104--10111},
    doi = {10.1073/pnas.1421398112}
}

@article{taylor1993artificial,
  title={Artificial life as a tool for biological inquiry},
  author={Taylor, Charles and Jefferson, David},
  journal={Artificial Life},
  volume={1},
  number={1\_2},
  pages={1--13},
  year={1993},
  doi={10.1162/artl.1993.1.1_2.1}
}

@article{taylor2016open,
  title={Open-ended evolution: Perspectives from the {OEE} workshop in {Y}ork},
  author={Taylor, Tim and Bedau, Mark and Channon, Alastair and Ackley, David and Banzhaf, Wolfgang and Beslon, Guillaume and Dolson, Emily and Froese, Tom and Hickinbotham, Simon and Ikegami, Takashi and McMullin, Barry and Packard, Norman and Agmon, Eran and Clark, Edward and McGregor, Simon and Ofria, Charles and Ropella, Glen and Spector, Lee and Stanley, Kenneth O. and Stanton, Adam and Timperley, Christopher and Vostinar, Anya and Wiser, Michael},
  journal={Artificial life},
  volume={22},
  number={3},
  pages={408--423},
  year={2016},
  publisher={MIT Press},
  doi={10.1162/ARTL_a_00210}
}

@article{taylor2019evolutionary,
    title = {Evolutionary innovations and where to find them: Routes to open-ended evolution in natural and artificial systems},
    author = {Tim Taylor},
    journal = {Artificial life},
    volume = {25},
    number = {2},
    pages = {207--224},
    year = {2019},
    doi = {10.1162/artl_a_00290}
}

@article{tennie2009ratcheting,
    title = {Ratcheting up the ratchet: on the evolution of cumulative culture},
    author = {Tennie, Claudio and Call, Josep and Tomasello, Michael},
    journal = {Philosophical Transactions of the Royal Society B: Biological Sciences},
    volume = {364},
    number = {1528},
    pages = {2405--2415},
    year = {2009},
    publisher = {The Royal Society},
    doi = {10.1098/rstb.2009.0052}
}

@article{tennie2018culture,
    title = {Culture, Cumulative},
    author = {Claudio Tennie and Christine Caldwell and Lewis G. Dean},
    year = {2018},
    journal = {The International Encyclopedia of Anthropology},
    doi = {10.1002/9781118924396.wbiea1998}
}

@article{tennie2020zone,
    title = {The zone of latent solutions and its relevance to understanding ape cultures},
    author = {Claudio Tennie and Elisa Bandini and Carel P. {van Schaik} and Lydia M. Hopper},
    journal = {Biology \& Philosophy},
    volume = {35},
    number = {5},
    pages = {1--42},
    year = {2020},
    publisher = {Springer},
    doi = {10.1007/s10539-020-09769-9}
}

@book{tomasello1999origins,
    title = {The Cultural Origins of Human Cognition},
    author = {Michael Tomasello},
    year = {1999},
    publisher = {Harvard University Press}
}

@article{uchiyama2021, 
    title={Cultural evolution of genetic heritability}, 
    doi ={10.1017/S0140525X21000893}, 
    journal={Behavioral and Brain Sciences}, 
    publisher={Cambridge University Press}, 
    author={Ryutaro Uchiyama and Rachel Spicer and Michael Muthukrishna},
    year={2021}, 
    pages={1–147}
}

@article{uhlivr2012needs,
  title={Who Needs Memetics? Possible Developments of the Meme Concept and Beyond},
  author={Uhl{\'{i}}{\v{r}}, Vil{\'e}m and Stella, Marco},
  journal={Anthropologie},
  volume={50},
  number={1},
  pages={127--142},
  year={2012},
  publisher={JSTOR},
}

@article{vanschaik2003orangculture,
    title = {Orangutan cultures and the evolution of material culture},
    author = {Carel P. {van Schaik} and Marc Ancrenaz and Gwendolyn Borgen and Birute Galdikas and Cheryl D. Knott and Ian Singleton and Akira Suzuki and Sri Suci Utami and Michelle Merrill},
    journal = {Science},
    volume = {299},
    number = {5603},
    pages = {102--105},
    year = {2003},
    doi = {10.1126/science.1078004}
}

@article{vanschaik2011social,
  title={Social learning and evolution: the cultural intelligence hypothesis},
  author={{van Schaik}, Carel P and Burkart, Judith M},
  journal={Philosophical Transactions of the Royal Society B: Biological Sciences},
  volume={366},
  number={1567},
  pages={1008--1016},
  year={2011},
  publisher={The Royal Society},
  doi={10.1098/rstb.2010.0304}
}

@article{waring2021longterm,
    title = {Long-term gene-culture coevolution and the human evolutionary transition},
    author = {Timothy M Waring and Zachary T Wood},
    journal = {Proceedings of the Royal Society B: Biological Sciences},
    year = {2021},
    pages = {20210538},
    volume = {282},
    doi = {10.1098/rspb.2021.0538}
}

@book{whitehead2015whales,
    title = {The Cultural Lives of Whales and Dolphins},
    author = {Hal Whitehead and Luke E Rendell},
    year = {2015},
    publisher = {University of Chicago Press}
}

@article{whitehead2019reach,
    title = {The reach of gene--culture coevolution in animals},
    author = {Hal Whitehead and Kevin N Laland and Luke Rendell and Rose Thorogood and Andrew Whiten},
    journal = {Nature Communications},
    volume = {10},
    number = {1},
    pages = {1--10},
    year = {2019},
    doi = {10.1038/s41467-019-10293-y}
}

@article{whiten1999cultures,
    title = {Cultures in chimpanzees},
    author = {Andrew Whiten and Jane Goodall and William C McGrew and T Nishida and V Reynolds and Y Sugiyama and C E Tutin and Richard W Wrangham and Christoph Boesch},
    journal = {Nature},
    year = {1999},
    pages = {6737},
    volume = {399},
    doi = {10.1038/21415}
}

@article{whiten2019cultevol,
    author = {Andrew Whiten},
    title = {Cultural evolution in animals},
    journal = {Annual Review of Ecology, Evolution, and Systematics},
    volume = {50},
    number = {1},
    pages = {27-48},
    year = {2019},
    doi = {10.1146/annurev-ecolsys-110218-025040}
}

@article{whiten2021burgeoning,
    title = {The burgeoning reach of animal culture},
    author = {Andrew Whiten},
    journal = {Science},
    volume = {372},
    number = {6537},
    year = {2021},
    doi = {10.1126/science.abe6514}
}

@article{whiten2021psychological,
    title = {The Psychological Reach of Culture in Animals’ Lives},
    author = {Andrew Whiten},
    journal = {Current Directions in Psychological Science},
    volume = {30},
    number = {3},
    pages = {211--217},
    year = {2021},
    doi = {10.1177/0963721421993119}
}

@article{whiten2022emergence,
  title={The emergence of collective knowledge and cumulative culture in animals, humans and machines},
  author={Whiten, Andrew and Biro, Dora and Bredeche, Nicolas and Garland, Ellen C and Kirby, Simon},
  journal={Philosophical Transactions of the Royal Society B},
  volume={377},
  number={1843},
  year={2022},
  publisher={The Royal Society},
  doi = {10.1098/rstb.2020.0306}
}

@article{winters2020comb,
    title={Is the cultural evolution of technology cumulative or combinatorial?},
    url={osf.io/preprints/socarxiv/aypnx},
    doi={10.31235/osf.io/aypnx},
    journal={SocArXiv},
    author={Winters, James},
    year={2020}
}

@book{lewens2015culture,
    title={Cultural Evolution},
    author={Lewens, Tim},
    publisher={Oxford University Press},
    year={2015}
}

@article{clay2017overimitation,
    title={Is Overimitation a Uniquely Human Phenomenon? {I}nsights From Human Children as Compared to Bonobos},
    author={Clay, Zanna and Tennie, Claudio},
    journal={Child Development},
    volume={89},
    issue={5},
    pages={1535--1544},
    year={2017}
}

@article{herrmann2007specialized,
    title={Humans Have Evolved Specialized Skills of Social Cognition: The Cultural Intelligence Hypothesis},
    author={Herrmann, Esther and Call, Joseph and Victoria Hern\'{a}ndez-Lloreda, Maria and Hare, Brian and Tomasello, Michael},
    journal={Science},
    volume={317},
    issue={1360},
    year={2007},
    doi={10.1126/science.1146282}
}

@article{stringer2016origin,
  title={The origin and evolution of Homo sapiens},
  author={Stringer, Chris},
  journal={Philosophical Transactions of the Royal Society B: Biological Sciences},
  volume={371},
  number={1698},
  pages={20150237},
  year={2016},
  publisher={The Royal Society},
  doi={https://doi.org/10.1098/rstb.2015.0237}
}

@article{stringer2017origin,
  title={On the origin of our species},
  author={Stringer, Chris and Galway-Witham, Julia},
  journal={Nature},
  volume={546},
  number={7657},
  pages={212--214},
  year={2017},
  publisher={Nature Publishing Group},
  doi={10.1038/546212a}
}

@article{lewis2016earlier,
  title={An earlier origin for stone tool making: implications for cognitive evolution and the transition to Homo},
  author={Lewis, Jason E and Harmand, Sonia},
  journal={Philosophical Transactions of the Royal Society B: Biological Sciences},
  volume={371},
  number={1698},
  pages={20150233},
  year={2016},
  publisher={The Royal Society},
  doi={10.1098/rstb.2015.0233}
}

@article{wagner1996perspective,
  title={Perspective: complex adaptations and the evolution of evolvability},
  author={Wagner, G{\"u}nter P and Altenberg, Lee},
  journal={Evolution},
  volume={50},
  number={3},
  pages={967--976},
  year={1996},
  publisher={Wiley Online Library},
  doi={10.1111/j.1558-5646.1996.tb02339.x}
}

@article{wilson1997altruism,
  title={Altruism and organism: Disentangling the themes of multilevel selection theory},
  author={Wilson, David Sloan},
  journal={The American Naturalist},
  volume={150},
  number={S1},
  pages={s122--S134},
  year={1997},
  publisher={The University of Chicago Press},
  doi={10.1086/286053}
}

@book{haldane1932causes,
  title={The causes of evolution},
  author={Haldane, John Burdon},
  volume={Reprinted 1990},
  year={1932},
  publisher={Princeton University Press}
}

\end{document}